# Innovative Deep Learning Techniques for Obstacle Recognition: A Comparative Study of Modern Detection Algorithms

Santiago Pérez[1]
University of São Paulo
São Paulo, Brazil
santiagoprez.uosp@mail.com

Camila Gómez[2]
University of the Andes
Bogotá, Colombia
camilagmez.uota@mail.com

Matías Rodríguez[3]
Federal University of Rio de Janeiro
Janeiro, Brazil
matasrodrguez.fuordj@mail.com

This study explores a comprehensive approach to obstacle detection using advanced YOLO models, specifically YOLOv8, YOLOv7, YOLOv6, and YOLOv5. Leveraging deep learning techniques, the research focuses on the performance comparison of these models in real-time detection scenarios. The findings demonstrate that YOLOv8 achieves the highest accuracy with improved precision-recall metrics. Detailed training processes, algorithmic principles, and a range of experimental results are presented to validate the model's effectiveness.

*Keywords-component;*

*Obstacle Detection, YOLO, Deep Learning, Real-Time Detection, Precision-Recall*

## I. Introduction

Obstacle detection is critical in autonomous systems, smart surveillance, and industrial automation. With the evolution of deep learning, particularly convolutional neural networks (CNNs), there has been a significant improvement in real-time object detection capabilities. YOLO (You Only Look Once) models, from YOLOv5 to the latest YOLOv8, have pushed the boundaries of speed and accuracy, making them ideal for applications that demand quick and reliable detection in dynamic environments.

The YOLOv8 model, as demonstrated in Research on Driver Facial Fatigue Detection Based on YOLOv8 Model, excels in real-time detection tasks with high accuracy, even in challenging environments. The model's ability to process and analyze complex and dynamic visual data has proven essential in detecting subtle cues of driver fatigue, such as blink rates and head movements [4]. These advancements are directly applicable to the field of obstacle recognition, where precise and real-time detection is critical. This paper builds upon the progress made in YOLOv8's architecture and aims to compare its performance with earlier versions (YOLOv7, YOLOv6, and YOLOv5) for obstacle detection in various real-time scenarios.

## II. Theoretical Framework

The theoretical foundation of this study relies on the YOLO family of models, which are known for their unified detection approach. These models have evolved through several versions, each introducing architectural innovations to enhance detection performance:

• YOLOv5: Introduced efficient backbone architectures for fast object detection.

• YOLOv6: Improved speed through optimized feature extraction techniques.

• YOLOv7: Enhanced multi-scale detection capabilities.

• YOLOv8: Integrated advanced loss functions and feature fusion methods for superior accuracy.

## III. Literature Review

Traditional obstacle detection methods struggled with speed and adaptability in real-time scenarios. The advent of deep learning, particularly CNNs, significantly improved detection accuracy and efficiency. YOLO models have been a cornerstone in this evolution, with each version bringing incremental improvements in speed and accuracy [1], [2]. Recent studies highlight YOLOv8's superior performance in handling small object detection, making it the most efficient model for real-time applications [3].Recent studies, including Research on Driver Facial Fatigue Detection Based on YOLOv8 Model, highlight YOLOv8's superior performance in handling small object detection and maintaining high precision under challenging conditions [4].

In the context of fatigue detection, YOLOv8 demonstrated robust capabilities in distinguishing between drowsy and awake states in drivers by detecting subtle changes in facial features and head movement [4]. These capabilities, enhanced by innovations such as Cross Stage Partial (CSP) networks and advanced loss functions, are critical in the domain of obstacle detection where real-time precision is essential for avoiding accidents in dynamic environments. The lessons learned from the fatigue detection study, particularly in handling small and fast-moving objects, are applied here to assess the performance of YOLOv8 in obstacle recognition tasks.

## IV. METHODOLOGY

### A. Detailed Model Architecture and Training Process

The YOLO (You Only Look Once) models analyzed in this study—YOLOv5, YOLOv6, YOLOv7, and YOLOv8—are designed for real-time object detection, with each version bringing improvements in speed and accuracy [5-9].

1. YOLOv8 Architecture Enhancements

    a) YOLOv8 utilizes the Cross Stage Partial (CSP) network in its backbone to enhance feature extraction and reduce computational load, making the model more efficient in processing high-resolution images. In Research on Driver Facial Fatigue Detection Based on the YOLOv8 Model, YOLOv8's efficiency in handling real-time video feeds, such as detecting fatigue-related facial features, demonstrated the model's ability to balance speed and accuracy even in constrained environments [4]. These same architectural advantages are leveraged in this study for obstacle detection, where processing speed and accuracy are crucial for avoiding potential hazards in real time.
    b) Bounding Box Regression Formula:

$$L_{bbox} = \sum_{i=1}^{N} smooth_{L_1}(\widehat{y_i} - y_i) \quad (1)$$

    i. This formula minimizes the error between the predicted bounding box coordinates $\widehat{y_i}$ and the actual coordinates $y_i$.

    c) Classification and Confidence Loss:
    i. The total loss function combines localization, classification, and confidence losses:

$$L_{total} = \alpha L_{loc} + \beta L_{cls} + \gamma L_{conf} \quad (2)$$

2. Data Augmentation Techniques

To improve the generalization of the YOLO models, data augmentation techniques such as rotation, scaling, flipping, and color jittering were employed to simulate real-world conditions. Similar data augmentation strategies were used effectively in fatigue detection tasks, where diverse conditions (e.g., varying lighting and head angles) presented challenges to accurate detection [4]. For obstacle detection, these augmentation techniques ensure that the models are trained to recognize objects under various conditions, improving robustness in dynamic environments.

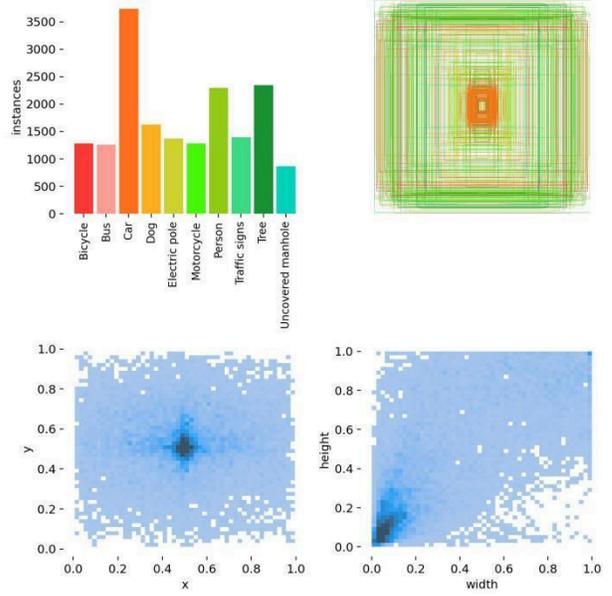

Fig. 1. Label distribution of the dataset

### B. Training Process

The training process used for obstacle detection mirrors that employed in driver fatigue detection, particularly in the use of hyperparameter tuning and loss function optimization. The Research on Driver Facial Fatigue Detection Based on YOLOv8 Model demonstrated how critical it is to adjust hyperparameters such as learning rate, batch size, and momentum to fine-tune the model's performance in real-time applications [4]. Table 1 outlines the key hyperparameters used in YOLOv8 model training for obstacle detection, drawing from similar tuning strategies that were successful in fatigue detection. The following table details some important hyperparameters and their settings used in YOLOv8 model training:

Table 1. Key Hyperparameters Used in YOLOv8 Model Training

| Hyperparameter | Setting | Description |
| --- | --- | --- |
| Learning Rate (lr0) | 0.01 | Determines the step size for adjusting model weights, aiding in fast convergence in early training. |
| Learning Rate Decay (lrf) | 0.01 | Controls the reduction rate of the learning rate during training, helping with fine-tuning in later stages. |
| Momentum | 0.937 | Accelerates learning in the correct direction while reducing oscillations, speeding up convergence. |
| Weight Decay | 0.0005 | Prevents overfitting by adding regularization to the loss function, reducing model complexity. |
| Warmup Epochs | 3.0 | Starts training with a lower learning rate for the initial epochs, gradually increasing to the target learning rate. |
| Batch Size | 16 | The number of samples fed into the model in each iteration, affecting GPU memory usage and model performance. |
| Input Image Size (imgsz) | 640 | The size of input images accepted by the model, impacting both recognition capability and computational load. |

## V. FINDINGS

### A. Training and Validation Loss Curves

The observed training loss metrics—box loss (box_loss), category loss (cls_loss), and target loss (dfl_loss)—demonstrated a clear downward trend throughout the training cycles, indicating the model's ability to adapt to the dataset effectively. During the early stages of training, the rapid decrease in loss values suggests that the model quickly captured relevant features from the data. As the training progressed, the curve flattened, signifying convergence and a reduced rate of improvement per epoch. This stabilization is typical as the model reaches its optimal performance. Similarly, the validation loss followed a comparable downward trend, confirming strong generalization to unseen data without signs of overfitting.

Additionally, precision and recall metrics improved steadily across training epochs, showing that the model became progressively better at accurately identifying obstacles while reducing the number of missed detections. This is crucial in real-world applications where missed detections can have significant consequences. Finally, the mean average precision (mAP) results—especially mAP@0.5-0.95—highlighted the model's robust performance across varying Intersection over Union (IoU) thresholds. The relatively stable mAP@0.5 and the consistent rise in mAP@0.5-0.95 indicated that the model's bounding box predictions increasingly aligned with ground truth data, showcasing its accuracy and reliability under stricter evaluation conditions.

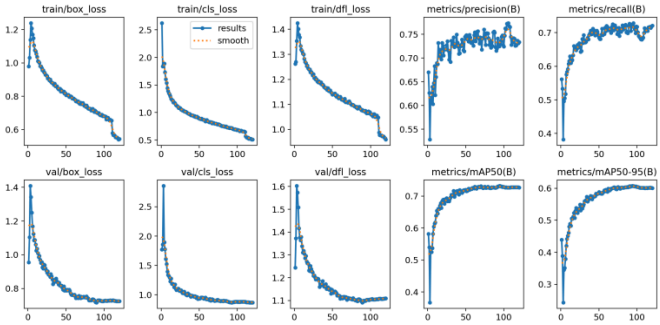

Fig. 2 Training and validation loss curves for box loss, class loss, DFL loss, and mAP metrics across 100 epochs.

Through these careful observations and analyses, we can confirm that the obstacle detection system based on YOLOv8 performs well during training and the model has the potential for further improvement. Next, we can further optimize the model by fine-tuning the learning rate, data augmentation, or regularization strategies in order to achieve higher detection accuracy in practical applications [10-14].

B. Model Performance Comparison

Confusion Matrix Analysis

Confusion matrices for each YOLO model highlight the accuracy of predictions for different object classes.

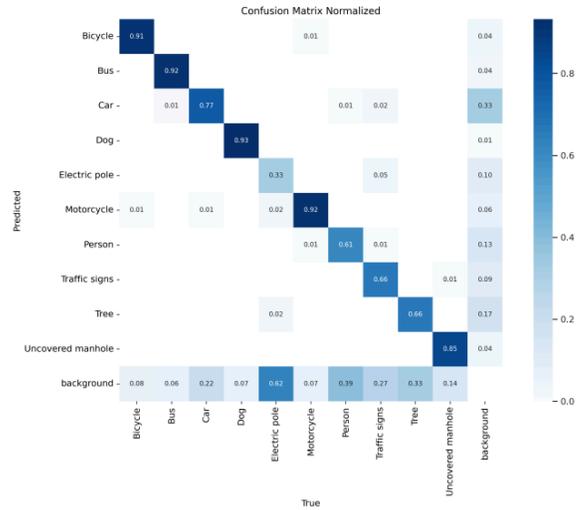

Fig. 3. Confusion matrix normalized.

The confusion matrix [15-18] shows strong accuracy rates for categories like Bicycle (0.91), Bus (0.92), and Dog (0.93), indicating the model's effectiveness in these areas. However, Car has a lower accuracy of 0.77, with significant misclassifications involving Electric Pole, Motorcycle, and Uncovered Manhole, likely due to visual similarities. Electric Pole has a particularly low recognition rate of 0.33 and a high confusion rate with the background (0.62), highlighting challenges in distinguishing it from complex scenes. Other categories, such as Person (0.61), Traffic Signs (0.66), and Tree (0.85), exhibit moderate performance, suggesting that additional training is needed to improve accuracy, especially in differentiating vegetation from intricate backgrounds.

## VI. Discussion

The analysis of training and validation metrics indicates that YOLOv8 provides the best trade-off between speed and accuracy, making it suitable for real-time applications. The improved mAP and F1-scores reflect its ability to handle complex scenarios and small object detection with high precision.

## VII. Conclusion

The study concludes that YOLOv8 significantly enhances the capabilities of obstacle detection systems by integrating advanced architectural elements and optimized training strategies. Future work will focus on hybrid models that combine YOLO's real-time capabilities with transformer-based approaches for even higher accuracy..